\documentclass[10pt,twocolumn,letterpaper,hyphens]{article}

\usepackage{cvpr}

\usepackage{graphicx}
\usepackage{amsmath}
\usepackage{amssymb}
\usepackage{booktabs}
\usepackage[textwidth=3.7cm]{todonotes}
\usepackage[pagebackref,breaklinks,colorlinks]{hyperref}

\usepackage{balance}

\usepackage[capitalize]{cleveref}

\crefname{section}{Sec.}{Secs.}
\Crefname{section}{Section}{Sections}
\Crefname{table}{Table}{Tables}
\crefname{table}{Tab.}{Tabs.}

\def\cvprPaperID{117} 
\def\confName{CVPR}
\def\confYear{2024}

\begin{document}

\title{NTIRE 2024 Challenge on Night Photography Rendering}
\author{
Egor Ershov
\and
Artyom Panshin
\and
Oleg Karasev
\and 
Sergey Korchagin
\and
Shepelev Lev
\and
Alexandr Startsev
\and
Daniil Vladimirov
\and
Ekaterina Zaychenkova
\and
Nikola Banić
\and
Dmitrii Iarchuk
\and
Maria Efimova
\and
Radu Timofte
\and
Arseniy Terekhin
\and
Shuwei Yue
\and
Yuyang Liu 
\and
Minchen Wei
\and
Lu Xu
\and
Chao Zhang
\and
Yasi Wang
\and
Furkan Kınlı
\and
Doğa Yılmaz
\and
Barış Özcan
\and
Furkan Kıraç
\and
Shuai Liu
\and
Jingyuan Xiao
\and
Chaoyu Feng
\and
Hao Wang
\and
Guangqi Shao
\and
Yuqian Zhang
\and
Yibin Huang
\and
Wei Luo
\and
Liming Wang
\and
Xiaotao Wang
\and
Lei Lei
\and
Simone Zini
\and
Claudio Rota
\and
Marco Buzzelli
\and
Simone Bianco
\and
Raimondo Schettini
\and
Jin Guo
\and
Tianli Liu
\and
Mohao Wu
\and
Ben Shao
\and
Qirui Yang
\and
Xianghui Li
\and
Qihua Cheng
\and
Fangpu Zhang
\and
Zhiqiang Xu
\and
Jingyu Yang
\and
Huanjing Yue
}


\maketitle

\begin{abstract}

This paper presents a review of the NTIRE 2024 challenge on night photography rendering.
The goal of the challenge was to find solutions that process raw camera images taken in nighttime conditions, and thereby produce a photo-quality output images in the standard RGB (sRGB) space. 
Unlike the previous year's competition, the challenge images were collected with a mobile phone and the speed of algorithms was also measured alongside the quality of their output. 
To evaluate the results, a sufficient number of viewers were asked to assess the visual quality of the proposed solutions, considering the subjective nature of the task.
There were 2 nominations: quality and efficiency.
Top~5 solutions in terms of output quality were sorted by evaluation time (see Fig. \ref{fig:scores_illustration}).
The top ranking participants' solutions effectively represent the state-of-the-art in nighttime photography rendering. 
More results can be found at \url{https://nightimaging.org}.

\end{abstract}

\section{Introduction}

In-camera processing is widely used to process raw images obtained directly from the sensor into photographies encoded in a standard color space, such as sRGB. 
The main objective of this processing is to produce images that are visually pleasing and that simultaneously realistically represent the captured scene. 
However, nighttime photography presents unique challenges that are not typically encountered in daytime photography. 
For example, while a single illuminant can often be assumed for daytime images, there are typically multiple illuminants present in nighttime scenes and these can be significantly different. 
This makes it difficult to determine which illuminant(s) should be primarily taken into account during scene color correction. 
Moreover, common photo-finishing strategies used for daytime images may not be appropriate for night images due to differences in lighting conditions. 

\begin{figure}[!t]
    \centering
    \includegraphics[width=\linewidth]{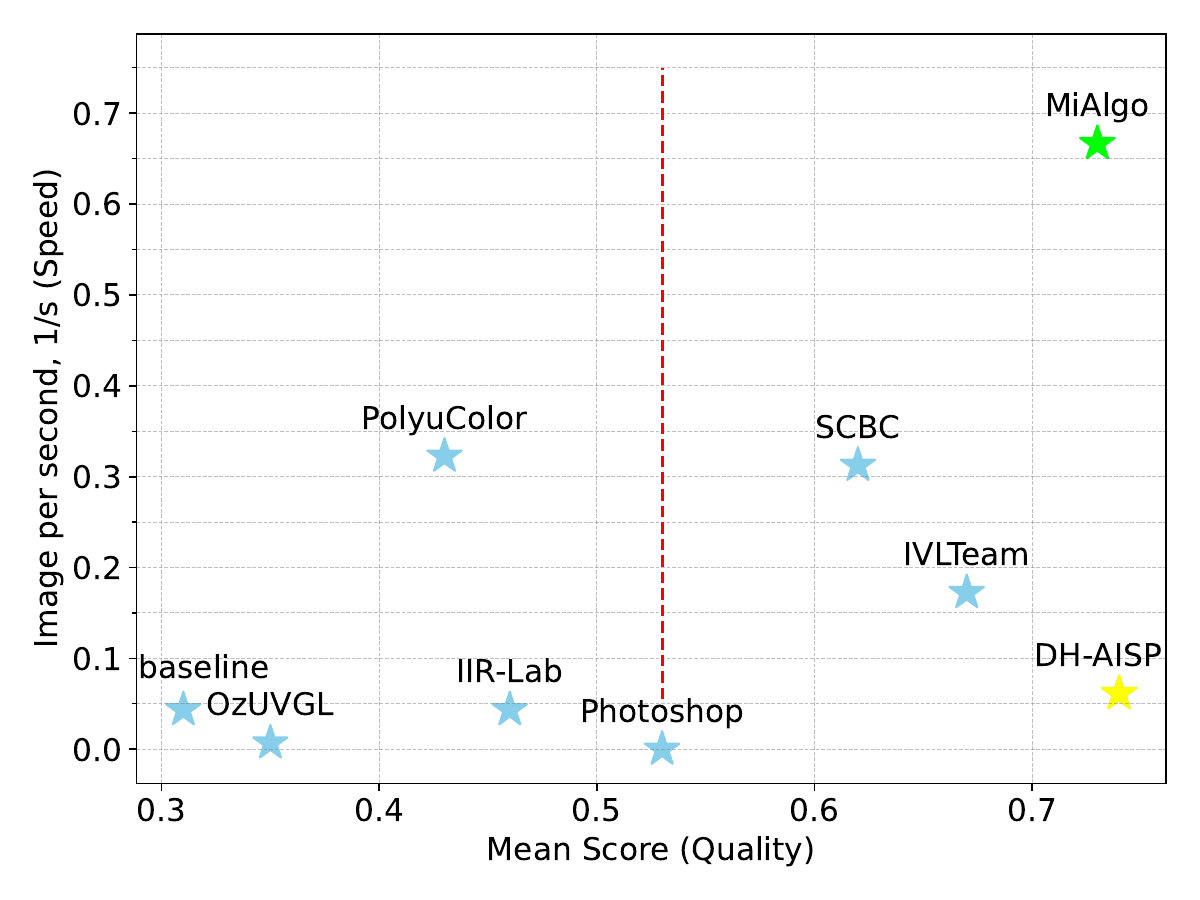}
    \caption{Illustration of final leader board scores and speed of algorithms. Red line separates top-5 quality solutions that were rearranged by inference speed. Winner of efficiency challenge is colored green, team that provided solution with best quality is colored yellow.}
    \label{fig:scores_illustration}
\end{figure}

Additionally, commonly used image metrics such as SSIM~\cite{wang2004image}, LPIPS~\cite{zhang2018unreasonable}, or MetaQA~\cite{zhu2020metaiqa}) do not appropriately assess the quality of night images. Furthermore, there is a dearth of published research focused specifically on image processing for night photography, resulting in fewer established ``best practices'' than for daytime photography. Having all this in mind, the main objective of this challenge is, similarly to the previous ones, to further encourage research into image processing techniques for night photography. The following sections provide a detailed description of the NTIRE challenge and the solutions proposed by the participating teams: Section~\ref{sec:challenge} lays out the challenge setup, Section~\ref{sec:results} describes the obtained results, Section~\ref{sec:discussion} presents a discussion on the obtained results, and Section~\ref{sec:teams} lists the teams with their members and their affiliations.

\begin{figure}[!t]
    \centering
    \includegraphics[width=\linewidth]{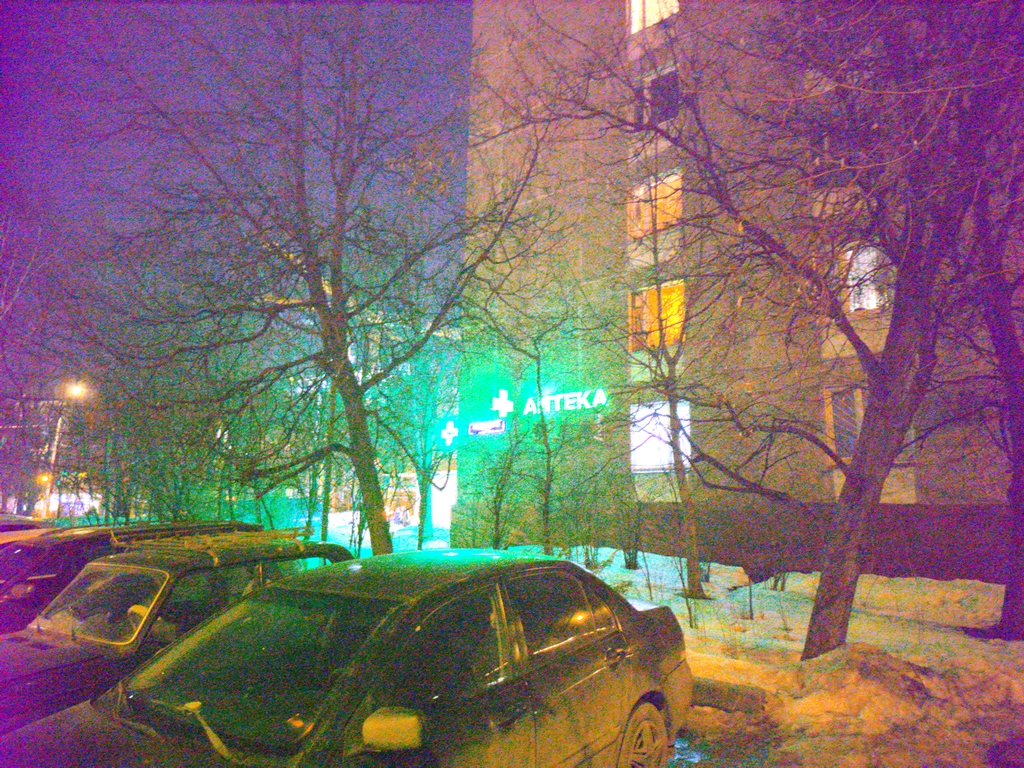}
    \caption{Example of the impact of high noise level and (color) vignetting on rendering a nighttime image when using only the baseline pipeline.}
    \label{fig:vignetting}
\end{figure}

\section{Challenge}
\label{sec:challenge}

In the challenge, the participating teams were required to develop automated solutions capable of producing visually appealing nighttime images. 
The challenge diverged from its predecessors as it incorporated raw images captured by mobile phones.
This shift is prompted by the prevalent demand for processing such images and the distinctive features they possess in contrast to the conventional camera images employed in prior challenges.

In addition to the already mentioned challenges, the competition participants faced the following ones as well:
\begin{enumerate}
    \item \textbf{High noise level.} Mobile phone camera pixel is $100$ times smaller than high-quality DSLR camera.
    \item \textbf{Vignetting and color vignetting.} Modern mobile lens systems not only suffer from traditional vignetting problems, but also experience significant color degradation from the center to the periphery.
    \item \textbf{Computational resources restriction.} Due to the limited computational resources on board of mobile phone, an extra ranking criteria has been introduced.
\end{enumerate}

The impact of the first and second challenge can be better understood by taking a look at Fig.~\ref{fig:vignetting} where the baseline pipeline has been used for image rendering and the mentioned problems were not taken into account.

As for the third challenge, the final ranking specifically prioritized the execution time of the top-performing solutions from the regular ranking, which is traditionally based solely on average quality as determined by the mean opinion score.

The teams' objective was to submit the corresponding rendered sRGB images obtained by the using the respective teams' solutions. 
Given the subjective nature of this task, the submissions were evaluated using mean opinion scores assigned by observers who were presented with pairs of two different renderings of the same scene and who then had to choose the rendering that they deemed visually more appealing.

This challenge is one of the NTIRE 2024 Workshop~\footnote{\url{https://cvlai.net/ntire/2024/}} associated challenges on: dense and non-homogeneous dehazing~\cite{ntire2024dehazing}, blind compressed image enhancement~\cite{ntire2024compressed}, shadow removal~\cite{ntire2024shadow}, efficient super resolution~\cite{ntire2024efficientsr}, image super resolution ($\times$4)~\cite{ntire2024srx4}, light field image super-resolution~\cite{ntire2024lightfield}, stereo image super-resolution~\cite{ntire2024stereosr}, HR depth from images of specular and transparent surfaces~\cite{ntire2024depth}, bracketing image restoration and enhancement~\cite{ntire2024bracketing}, portrait quality assessment~\cite{ntire2024QA_portrait}, quality assessment for AI-generated content~\cite{ntire2024QA_AI}, restore any image model (RAIM) in the wild~\cite{ntire2024raim}, RAW image super-resolution~\cite{ntire2024rawsr}, short-form UGC video quality assessment~\cite{ntire2024QA_UGC}, low light enhancement~\cite{ntire2024lowlight}, and RAW burst alignment and ISP challenge.

\subsection{Challenge Data}
The raw images of night outdoor and low-light indoor scenes were captured using Huawei Mate 40 Pro and encoded in 16-bit PNG files with additional meta-data provided in JSON files.
The challenge started with an initial 200 images provided to participants for algorithm development and testing. 
Three validation datasets with 125 images each were provided during the challenge.
Moreover, the calibration white low-light scenes were provided to participants as well.
A baseline code was provided to emulate the basic in-camera rendering as a starting point.

The majority of the images were taken in the urban area and this category can be divided into images of open and closed spaces. 
There are also indoor images.
These can all be designated as categories 1, 2, and 3. 
The first validation dataset contains more images of open spaces, with a ratio of images between these categories being 100/25/0. 
The second validation dataset already includes indoor images, with a ratio of 108/6/11. 
The third dataset has an even greater number of indoor images, with a ratio of 89/3/33.

The final dataset is focused on category 1 images, which dominate the data, but it includes other categories as well. 
The ratio of categories in it is 39/7/4.
The dataset was made publicly available.\footnote{\url{https://zenodo.org/records/10931755}}

\subsection{Evaluation}

The evaluation consisted of three validation checkpoints during the contest and a final evaluation to determine the winners. 
Mean opinion scores were obtained using Yandex Tasks (a service similar to Mechanical Turk) for the checkpoints and final evaluation. 
Yandex Tasks users ranked their preferred solutions in a forced-choice manner with a question: ``Which image is more pleasant?''.
The answer options were: ``left'', ``right'' or ``they are the same''.
To ensure basic quality control, all Yandex Tasks users who chose ``left'' of ``right'' for a pair of same images have been banned, while all their previous answers have been declined. 
It is worth noting that in our setup Yandex Tasks mainly relied on users from Eastern Europe to perform the image ranking. 
As a result, there may be a cultural bias in terms of the preferred image aesthetics by the users. 
All solutions have been anonymized to guarantee unbiased results.

As recently shown~\cite{visapp24}, the Yandex Tasks platform generates stable and reproducible results. Also, we chose only 10\% best voters and conducted filtering of votes in order to exclude any fraud scenarios.

As we implement the pairwise approach for our experiments, the raw data obtained for the pairwise comparisons of the style renderings, i.e., versions of a single image can be denoted as $A_{ijt}$ where $i$ and $j$ denote the $i$-th and $j$-th image versions, i.e., team renderings that were shown to the voters and $t$ enumerates the participants who evaluated the $(i, j)$ image pair.
$A_{ijt}$ is equal to $1$ if the $t$-th participant evaluated the $i$-th image version to be better than the $j$-th and $0$ otherwise.
Then, the score for an image is calculated as $S_i=\frac{1}{NT}\sum_{j=1}^N\sum_{t=1}^TA_{ijt}$, where $N$ is total number of evaluated solutions and $T$ is total number of voters.
The final score for team's solution is obtained as $S = \frac{1}{N}\sum_{i=1}^NS_i$.

To measure speed, all submitted solutions were executed on the same computer with the following technical specifications:
\begin{itemize}
    \item \textbf{CPU}: Intel(R) Core(TM) i7-4790 CPU @ 3.60GHz
    \item \textbf{RAM}: 16.0 GB
    \item \textbf{GPU}: MSI GeForce RTX 2060 12Gb
\end{itemize}
Only the actual image processing time was measured, excluding image loading and saving.

During each validation checkpoint, 125 new test images were provided, and each participating team was allowed to submit up to two distinct solution image sets, each consisting of exactly 125 images. 
The purpose of having three validation sets was to allow participants to test different solutions' behavior and receive feedback on their solution's quality.

For the final submission, only one solution was allowed, and 50 hidden test images were used for the final validation.  
The user study images were generated using the code provided by the participants by means of Docker. 
Only open and reproducible results were accepted. 
The top 5 solutions according to Yandex Tasks were further sorted by performance speed.

\begin{figure}[h]
    \centering
    \includegraphics[width=1.0\linewidth]{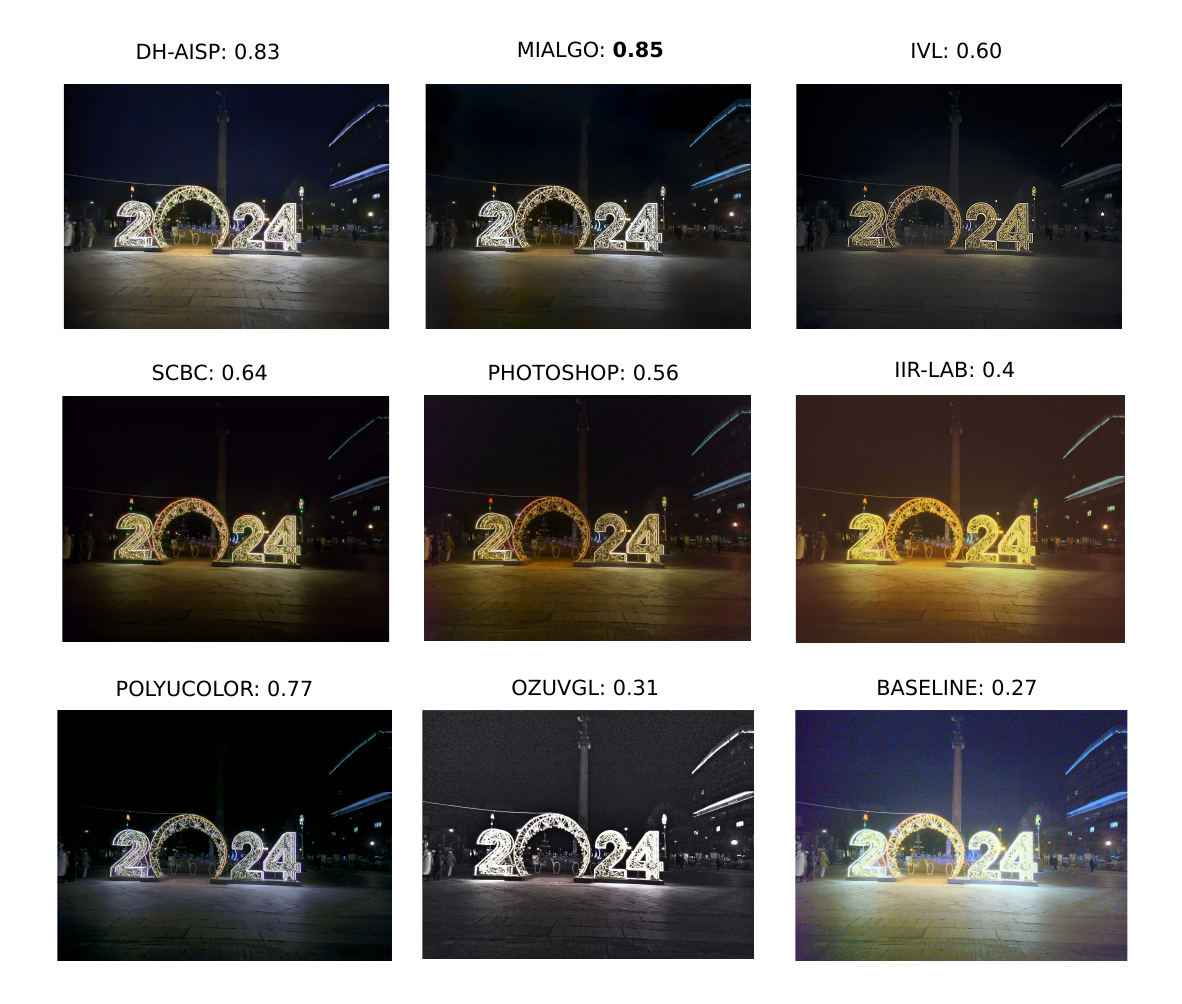}
    \caption{One scene from the final validation set and all results provided by participants. The best quality image is in bold.}
    \label{fig:submissions_illustartion}
\end{figure}

In the context of self-assessment by people, algorithms from the MIALGO and DH-AISP teams proved to be the best.
The downside of the second solution is the excessive lighting of dark images, as well as a very long processing time, namely 11 times longer than the first algorithm.
An example of final solutions' images is presented in Fig.~\ref{fig:submissions_illustartion}.

\section{Results}
\label{sec:results}

The section presents the ranking results obtained using the Yandex Tasks service and performance evaluation.

\subsection{People's Choice and Discussion}

Table~\ref{tab:people} provides the ranking of the mean opinion reported by Yandex Tasks users for the different teams' final submissions (quality challenge). 
Table~\ref{tab:efficciency} provides ranking for efficiency challenge.
Table~\ref{tab:efficciency} is made up from Table~\ref{tab:people} with sorting top 5 solutions by speed performance.

\begin{table}[h!]
    \centering
    \begin{tabular}{lllll}
        \hline
        Rank & Team & Mean Score & Time, s \\
        \hline
        \hline
        1 & DH-AISP & 0.74 & 16.3\\
        2 & MiAlgo & 0.73 & 1.5 \\
        3 & IVLTeam & 0.67 & 5.8 \\
        4 & SCBC & 0.62 & 3.2 \\
        5 & Manual enhancement & 0.53 & $\infty$ \\
        \hline
        6 & IIR-Lab & 0.46 & 23.0 \\
        7 & PolyuColor & 0.43 & 3.1 \\
        8 & OzUVGL & 0.35 & 144.8 \\
        9 & baseline & 0.31 & 23.0  \\
        \hline
    \end{tabular}
    \caption{People's choice ranking results. Quality leaderboard.}
    \label{tab:people}
\end{table}
\begin{table}[h!]
    \centering
    \begin{tabular}{lllll}
        \hline
        Rank & Team & Mean Score & Time, s \\
        \hline
        \hline
        1 & MiAlgo & 0.73 & 1.5 \\
        2 & SCBC & 0.62 & 3.2\\
        3 & IVLTeam & 0.67 & 5.8 \\
        4 & DH-AISP & 0.74 & 16.3\\
        5 & Manual enhancement & 0.53 & $\infty$ \\
    \end{tabular}
    
    \caption{Top 5 solutions are sorted by inference time. Efficiency leaderboard.}
    \label{tab:efficciency}
\end{table}
This year's competitors have presented a diverse range of solutions that produce visually appealing images.

\subsection{Teams' solutions}

This section contains brief description of the submitted solutions by participants. More detailed description and code will be available on the challenge repository \footnote{\href{https://github.com/createcolor/nightimaging24}{https://github.com/createcolor/nightimaging24}}. 

\subsubsection{Baseline}

In this year's challenge, two baseline methods were given to the participants to use: a simple classic ISP and manual image enhancement. 
The simple classic pipeline involved debayering using linear interpolation, white balancing by using the Gray World method~\cite{buchsbaum1980spatial}, a mean throughout train dataset matrix for CST, and a standard transform from XYZ to sRGB. 
This pipeline was also provided as a baseline for the participants\footnote{available at challenge repository}.

To enhance the images manually, we employed the Adobe Camera RAW application and invited non-professional photographers to participate. Each image was corrected individually within a short span of 3 to 5 minutes. The corrections comprised of adjusting the temperature to cool down the image, adding a violet tint, increasing the brightness via exposure adjustment, enhancing the contrast, reducing the highlights, brightening the shadows, and reducing the whites. Finally, the built-in noise reduction and color mixer were used to correct the hue and intensity of red, orange, and yellow (and sometimes blue and purple).

\subsubsection{MiAlgo}

\begin{figure}
    \centering
    \includegraphics[width=\linewidth]{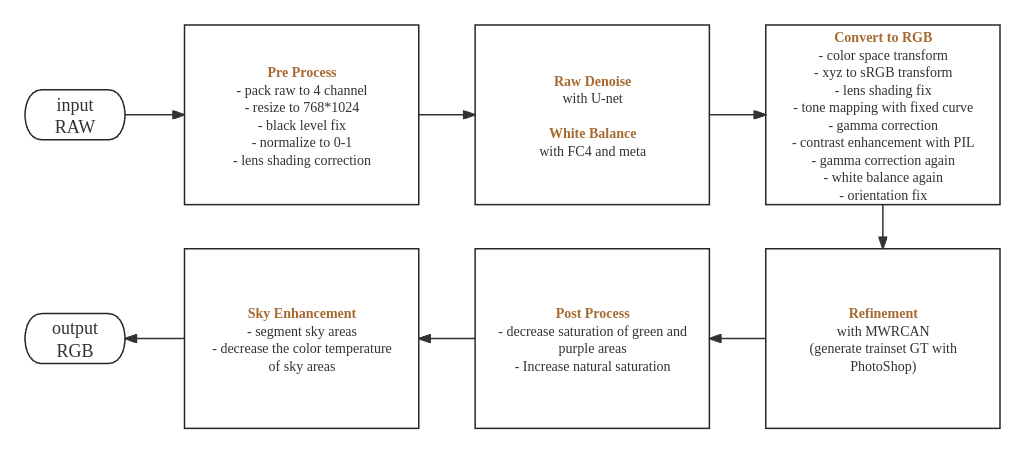}
    \caption{Overall pipeline of MiAlgo team.}
    \label{fig:MiAlgo}
\end{figure}

We made improvements based on Deep-FlexISP\cite{liu2022deep}, and the overall pipeline is shown in Figure~\ref{fig:MiAlgo}. 

First, we pre-process the input image, including packing raw to 4 channels, down-sampling the image (to $768 \times 1024$), correcting the black level (adjusting value based on the noise profile), normalizing (to 0-1), and correcting lens shading (adjusting shading mask based on the provided calibration image and the noise profile). It is worth noting that we downsample the image to the required size at the beginning, which ensures that our whole processing is as fast as possible.

Next, we use the pre-trained Unet\cite{ronneberger2015u, liu2022deep} for raw domain denoising, and use the pre-trained FC4\cite{hu2017fc4} and meta info (as\_shot\_neutral) for white balance parameters estimation. 

Then we convert the raw image to RGB image with color space transform (fixed matrix), XYZ to sRGB transform (fixed matrix), lens shading fix (increase shading of dark scenes and decrease saturation), tone mapping (fixed curve), gamma correction (fixed parameter), contrast enhancement (python-pillow-autocontrast\cite{Pillow}), gamma correction again (fixed parameter), white balance correction again (grayness index\cite{qian2019finding}), and orientation fix. 

We use neural networks for the refinement of the RGB images. The model structure is based on MWRCAN\cite{ignatov2020aim}. The ground truth of the training data is generated using Photoshop. 

Then we post-process the image, decreasing the saturation of green and purple areas (uncommon and weird colors at night), and increasing the natural saturation of the whole image. 

Finally, we train a segmentation model\cite{wan2023seaformer, mmseg2020} to segment the sky areas and decrease the color temperature, which makes the sky more blue and cold.

\subsubsection{SCBC}

\begin{figure}
    \centering
    \includegraphics[width=0.9\linewidth]{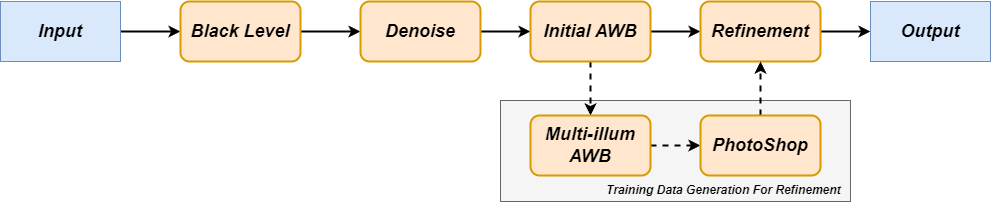}
    \caption{SCBC team pipeline scheme.}
    \label{fig:SCBC}
\end{figure}

SCBC team proposed a multi-stage ISP pipeline as shown in Fig.~\ref{fig:SCBC}, including pre-processing, Raw Image Denoising and Image refinement with Color Correction.
For raw image denoising, pretrained models are used \cite{abdelhamed2020ntire}. After having the initial images, we process the images with a multi-illumination color constancy method to solve the complex white balance introduced by artificial light source. The model is trained on RGB image with multi-illum lighting condition \cite{kim2021large}. The result are further manually adjusted and serve as the target image for refinement network training.  

\subsubsection{IVLTeam}

\begin{figure}
    \centering
    \includegraphics[width=\linewidth]{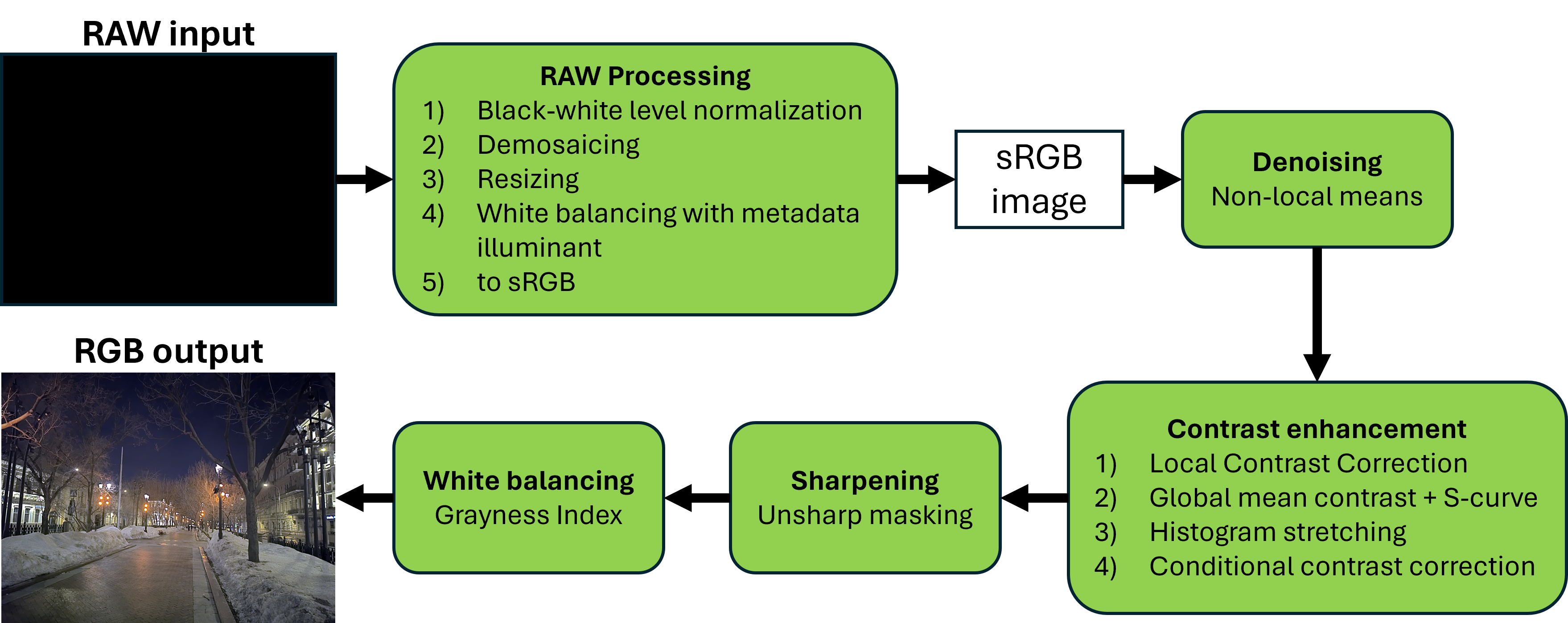}
    \caption{IVLTeam pipeline schema.}
    \label{fig:ivlteam}
\end{figure}

The solution proposed by IVLTeam is illustrated in Fig.~\ref{fig:ivlteam}. It relies on conventional image processing techniques and consists of five stages. It is based on previous works~\cite{zini2023shallow,zini2023back}, where the parameters of several processing steps are adapted to the challenge images and requirements.

The \textbf{first stage} works in the RAW domain and consists of five steps: black and white levels image normalization, raw demosaicing, image resizing to the target resolution, white balancing using the illuminant provided in the metadata, and conversion from the camera-sensor color space to the sRGB color space.

The \textbf{second stage} consists of a denoising operation using the Non-local means algorithm~\cite{buades2005non}. The intensity of denoising is proportional to the noise standard deviation estimated in the image using the method in~\cite{donoho1994ideal}. Stronger denoising is applied to the color channels than to the luma channel to effectively remove color noise while preserving image details and edges.

The \textbf{third stage} is a sequence of algorithms that enhance image contrast by manipulating the histogram distribution. First, the Local Contrast Correction (LCC) algorithm in~\cite{moroney2000local} is applied. 
As this process tends to decrease the overall contrast and saturation, the next step consists of a contrast and saturation enhancement using the approach proposed in~\cite{schettini2010contrast}.
Then, three steps to improve the image color appearance are applied. The first one adjusts the contrast by stretching the pixel values by a $\beta$ factor around their mean.
The second one is the application of the S-curve defined in~\cite{kang2010personalization}, where the center of the curve is set to zero, transforming the function into a gamma-like operation. The third one consists of a histogram stretching operation that increases the dynamic range and improves the overall contrast. After these operations, an extra conditional contrast correction operation, consisting of an additional S-curve or gamma correction, is applied depending on the mean value of the histogram. This improves visibility for very dark images, and restores the mood of nighttime scenes when the processed image is too bright.

The \textbf{fourth} and \textbf{fifth stages} perform sharpening and white balancing, respectively. Unsharp masking is used to enhance image details, which may have been flattened by the denoising operation in the second stage. White balancing is performed using the Grayness Index (GI) algorithm~\cite{qian2019finding} to further reduce color casts. 
As GI is sensitive to noise, the illuminant is estimated on a blurred version of the image and later applied to the sharp version.


\subsubsection{DH-AISP}

\begin{figure}
    \centering
    \includegraphics[width=\linewidth]{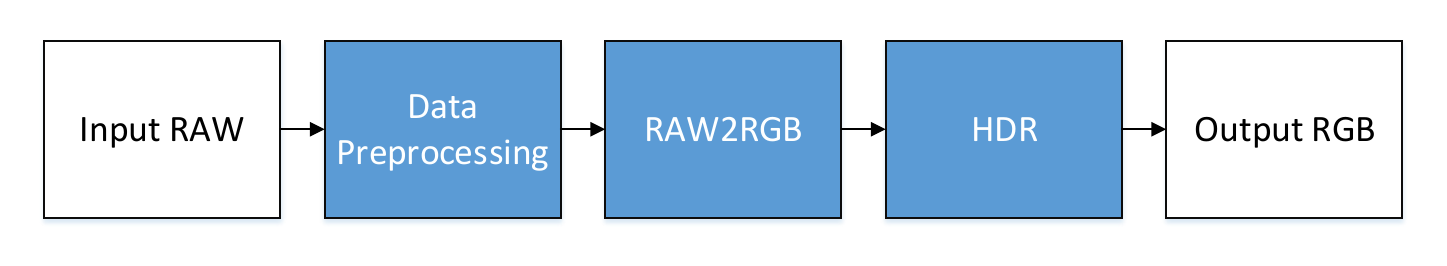}
    \caption{NISP-net}
    \label{fig:DH-AISP}
\end{figure}

Our main goal is to develop a technology for creating realistic and visually pleasing photographs of night scenes. By considering the data quality and modality, we outline the network structure in Fig.~\ref{fig:DH-AISP}. It contains three parts, involving the data pre-processing module, RAW to RGB module, and HDR module, which are elaborated in the following parts.

\noindent\textbf{Data pre-processing.} This module contains four steps: black-and-white balance correction, automatic white balancing, adaptive gain adjustment, and binning. The adaptive gain adjustment module generates a set of underexposed, normal exposed, and overexposed images by calculating the mean of the raw data. Binning is a method that sacrifices resolution to achieve higher signal-to-noise ratio. With these modules, we can obtain a corrected RAW data, which is a necessary step to achieve normal and better results.

\noindent\textbf{RAW to RGB.} We have trained a U-Net\cite{ronneberger2015u} structure to obtain images under different exposure gains.

\noindent\textbf{HDR.} Night images often have multiple illuminants, and light distribution of contents varies with the locations. Consequently, learning a unified light distribution in the existing methods produce undesired results. In this way, we propose to a controllable parameter to adjust the output of raw2rgb module, which allows us to produce the underexposed and overexposed candidates with different light distribution. Following that, an exposure fusion model based on U-Net to adaptively learn the fusion weights, and generates an image with satisfactory overall brightness. Finally, the CCM algorithm is introduced to further optimize the color distribution of the final output.

\subsubsection{IIR-Lab}
\begin{figure}
    \centering
    \includegraphics[width=\linewidth]{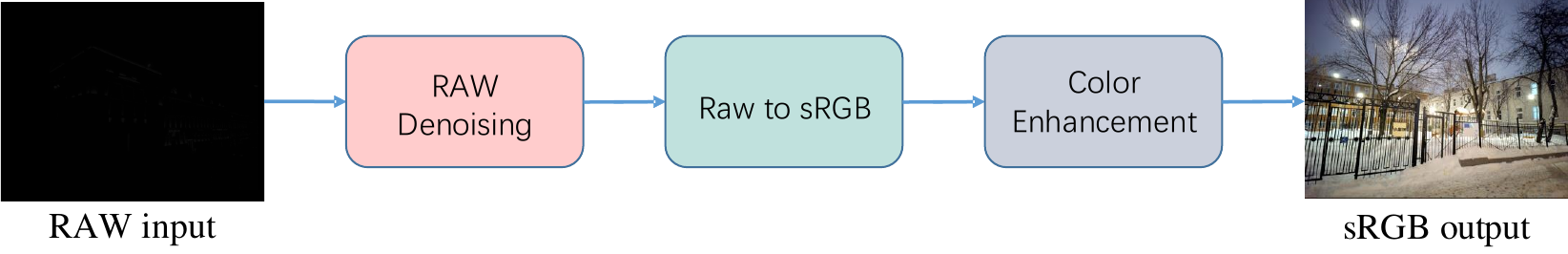}
    \caption{Illustration of our proposed ISP framework.}
    \label{fig:bigguy}
\end{figure}

Our main goal is to develop a technology for creating realistic and visually pleasing photographs of night scenes. By considering the data quality and requirements, we construct a new ISP pipeline, outlined in Fig.~\ref{fig:bigguy}. It contains three parts, involving the RAW denoising module, raw to RGB module, and color enhancement module, which are elaborated in the following parts.

\noindent\textbf{RAW denoising.} To capture more details with better visual effects, a higher gain is commonly used for the sensor, but simple gain multiplication causes noise degradation. To solve this problem, we first processed the Bayer domain denoising module, which utilizes NAF \cite{Chen_Chu_Zhang_Sun_2022}. 

\noindent\textbf{RAW to RGB.} Since the provided json file contains information such as AWB and CCM, this part of our implementation is done using traditional methods.

\noindent\textbf{Color Enhancement.} 
Night images often have complex noise, and multiple illuminants, and light distribution of contents varies with the locations. Therefore, we collect a set of nighttime images containing different light sources and render the corresponding ground truth using PS tools. We propose a UNet-based rendering network that enables real-time rendering on most devices. The final results show that our rendering network replaces the traditional modules of local and global Tonemapping, color enhancement, and sharpening.

\subsubsection{PolyuColor}
\begin{figure}[t]
    \centering
    \includegraphics[width=\linewidth]{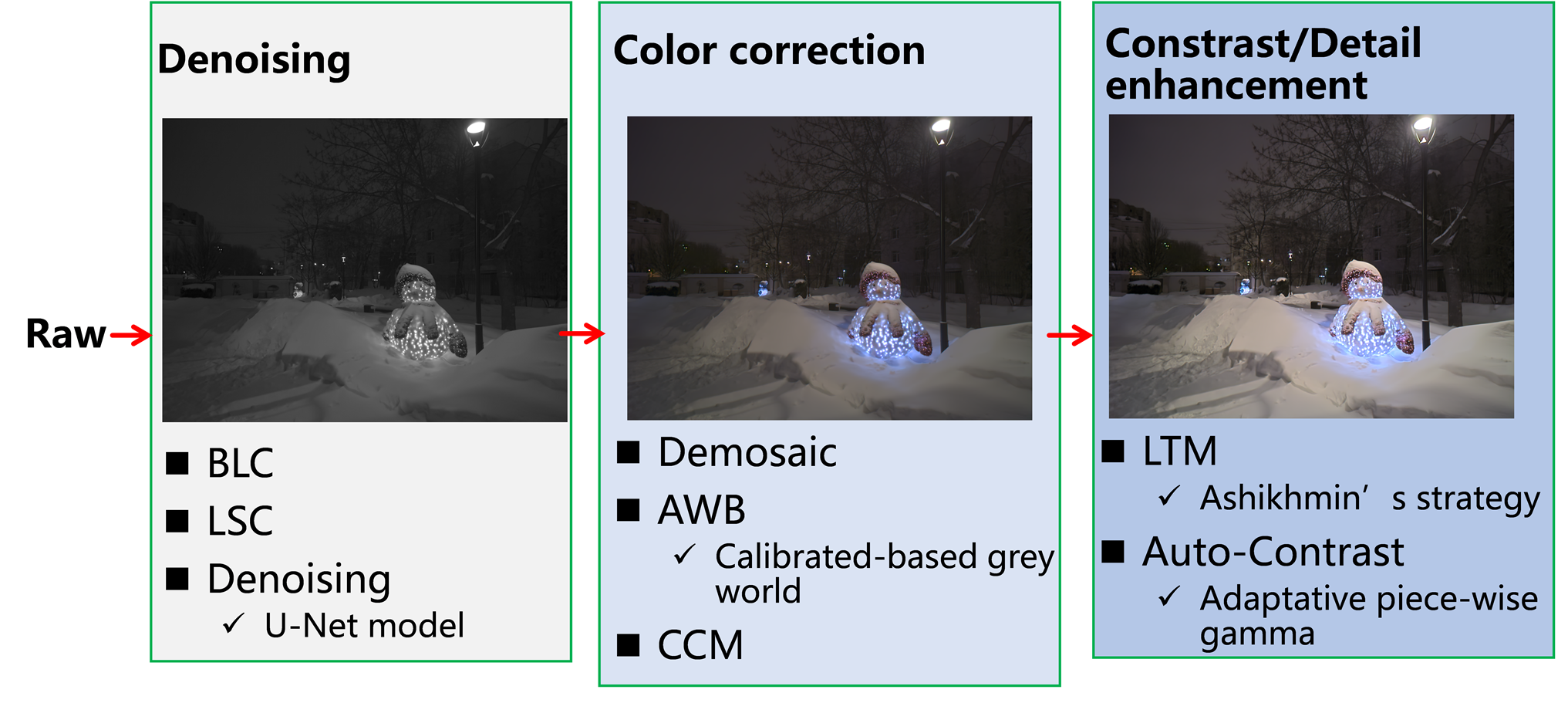}
    \caption{PolyuColor team pipeline schema.}
    \label{fig:PolyuColor}
\end{figure}



Nighttime imaging is challenging due to its low signal-to-noise ratio (SNR) and complicated lighting conditions. However, the three key elements that significantly impact image quality include denoising, color correction, and contrast/detail enhancement. In our method, a DNN-based approach is used for denoising, and traditional ISP algorithms are used for the other tasks, as shown in Fig.~\ref{fig:PolyuColor}. We aim to maintain superior image quality while minimizing computational resource consumption.

\noindent \textbf{Denoising.}~For RAW images, we initially apply black/white level (BLC) normalization and lens shading correction (LSC) as outlined in the baseline \cite{nightimaging2022}. Following this, a U-Net-based~\cite{chen2018learning} framework is employed to develop a model trained on the SID dataset~\cite{chen2018learning}, with the long-exposure images serving as the ground truth, supplemented by a synthetic noise model for comprehensive training.

\noindent \textbf{Color correction.}~We follow the methods used by the baseline for demosaicing and the color correction matrix (CCM). For auto white balance (AWB), our approach utilizes a calibrated-based grey world method~\cite{buchsbaum1980spatial}. By manually calibrating a set of white points and clustering them to approximate the sources on the Planckian locus, we define the correlated color temperature range within a circle, which is similar to the method in~\cite{Yue:23, Yue:24}. The image is segmented into several patches, with the grey world assumption adopted to derive the white point for each patch. The final white point is the weighted average of the cluster centers based on the distance between the white point of each patch and the corresponding cluster center(s).

\noindent \textbf{Contrast/Detail enhancement.}~After obtaining the sRGB images, we apply local tone mapping (LTM) utilizing Ashikhmin's method~\cite{ashikhmin02}. 
A scaling operation is performed to enhance the saturation by applying unequal gains on the three channels, which is different from the original method. An adaptive piece-wise gamma adjustment is finally applied to further enhance the global contrast.

\subsubsection{OzU-VGL}
\begin{figure}
    \centering
    \includegraphics[width=\linewidth]{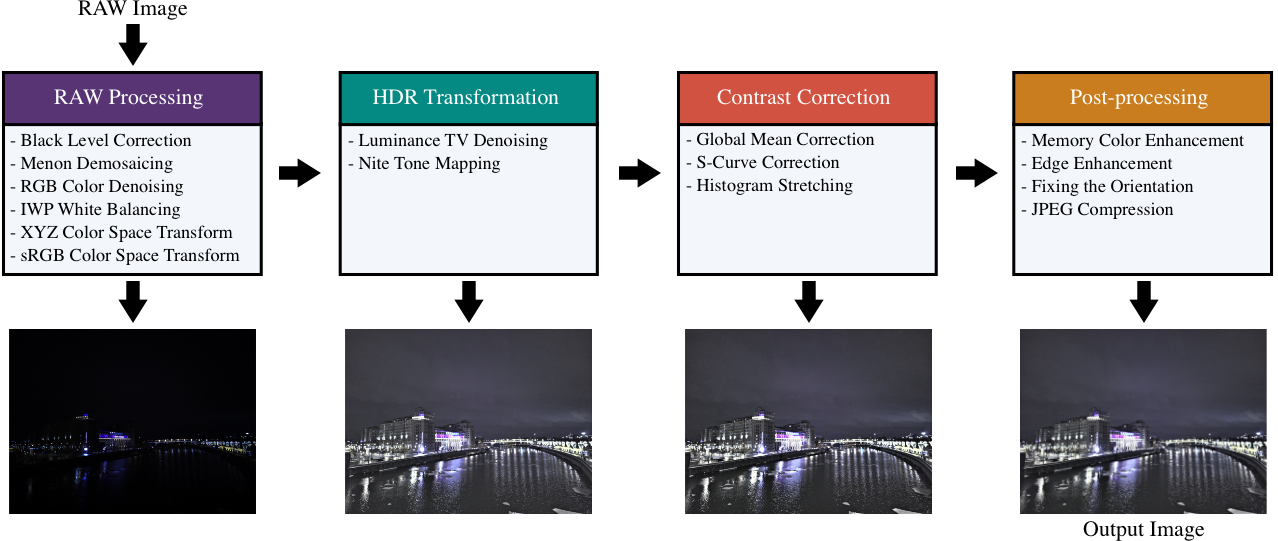}
    \caption{OzU-VGL pipeline.}
    \label{fig:vgl-ozu}
\end{figure}
In our image signal processor (ISP) pipeline for this challenge, we mainly focus on reproducing the previous year's winner solution~\cite{zini2023back} and improving it by including the improved version of the current state-of-the-art tone mapping operators (\textit{i.e.}, Flash and Storm \cite{banic2018flash}) where image statistics dynamically adjusts their scale parameter. Moreover, we applied simple tricks to avoid color casting in bright regions and the noise in dark regions in the final output. Fig.~\ref{fig:vgl-ozu} presents the proposed ISP pipeline for this challenge.

\noindent\textbf{RAW Processing.} The RAW input images are given in PNG format as the data, and we first applied black-level correction to the input by using the values in the metadata. We utilized the Directional Filtering algorithm \cite{Menon2007DemosaicingWD}, also referred to as \textit{Menon}, in our pipeline for demosaicing the RAW input, offering a more sophisticated alternative to the default CFA interpolation. Removing the noise on color channels in images with low grayscale mean value can be critical for avoiding undesired color casting on bright regions, so we applied simple Gaussian denoising to the color channels in YCbCr color space. Then, we integrated the random subsampling-based White Patch algorithm \cite{banic2014improving} into our pipeline to mitigate the yellowish effect induced by the Gray World algorithm, particularly evident in images with low grayscale mean values. We have implemented thresholds for the first and third diagonal values of the illuminant estimation matrix to avoid the occurrence of a greenish tint in bright areas caused by randomness. It is worth noting that, due to the randomness, it may be hard to reproduce the exact outputs for consecutive runs. We have submitted illuminance estimation matrices found by the final run during the final test to ensure reproducibility. The subsequent procedure involves converting the image from the raw-RGB to the sRGB. This is achieved by initially transforming raw-RGB to the XYZ color space, then converting XYZ to the sRGB using the specified color matrix tailored to Huawei Mate 40 Pro. At this point, we point out that some images produce salt-paper-like noise in dark regions, dramatically amplified by any tone mapping operator in the final output. 

\noindent\textbf{HDR Transformation.} To remove this noise factor, we applied total variation denoising \cite{getreuer2012rudin} to the luminance channel in YCbCr color space before applying HDR transformation. Apart from \cite{zini2023back}, we refrained from using local or conditional contrast correction operators. Instead, we introduced a new tone mapping technique named \textit{Nite}, specifically designed to enhance Storm's performance in low-light conditions. The operator employs adaptive adjustment of the scaling parameter $\alpha$ within the Naka-Rushton equation, utilizing image statistics. 

\noindent\textbf{Contrast correction.} To adjust the overall image color appearance, we applied three operations: (1) the global mean contrast operation enhances image contrast by scaling the values of RGB channels around their respective mean. (2) The S-curve correction applies to each RGB channel the S-curve proposed in \cite{kang2010personalization}. (3) The histogram stretching operation stretches the image histogram, thereby augmenting the dynamic range and enhancing overall contrast.

\noindent\textbf{Post-processing.} The memory color enhancement algorithm \cite{Braun2006MemoryCE} is employed to balance the colors of the sky, grass, and other specific hues in the contrast-corrected images. Subsequently, unsharp masking is applied to enhance edge sharpness, followed by aligning the image orientations based on metadata. Finally, the results are resized to the specified dimensions (\textit{i.e.}, $1024 \times 768$) and saved in JPG format to produce the final sRGB output.


\section{Discussion}
\label{sec:discussion}

A notable trend from this year's competition is that three out of the top five solutions, including the winning one, were enhancements of the solutions from prior years. This highlights the value of iterative improvement and underscores a continuous progression in this challenging field.

Importantly, the efficiency-winning solution of this year not only outperformed its top five peers in speed but was also the fastest overall.
In terms of image quality, it ranked second, narrowly trailing behind the top contender, which represents a desirable combination of qualities for use in production.

Furthermore, it's intriguing to observe that two solutions employing deep learning techniques surpassed a traditional method in computational efficiency.
This may denote a shift towards the readiness of employing such deep learning models in image processing pipelines of mobile phones.

\section{Teams and Affiliations}
\label{sec:teams}

\noindent \textit{\textbf{Team:}}\\
Organizers\\
\textit{\textbf{Members:}}\\
Artyom Panshin$^1$, Egor Ershov$^2$, Maria Efimova$^1$, Dima Yarchuk$^1$, Nikola Banić$^3$, Radu Timofte$^4$, Oleg Karasev$^1$, Sergey Korchagin$^1$, Arseniy Terekhin$^1$, Alexandr Startsev$^1$, Daniil Vladimirov$^1$, Ekaterina Zaychenkova$^1$, Shepelev Lev$^1$  \\
\textit{\textbf{Affiliations:}}\\
$^1$:~Institute for Information Transmission Problems (Kharkevich institute), Russia\\
$^2$:~Artifitial Intelligence Research Institute (AIRI), Russia\\
$^3$:~Gideon Brothers, Croatia\\
$^4$:~Swiss Federal Institute of Technology in Zurich, Switzerland and University of Würzburg, Germany\\

\noindent \textit{\textbf{Team:}}\\
PolyuColor\\
\textit{\textbf{Members:}}\\
Shuwei Yue$^1$, Yuyang Liu$^1$, Minchen Wei$^{1,2}$\\
\textit{\textbf{Affiliations:}}\\
$^1$: Color, Imaging, and Illumination Laboratory, The Hong Kong Polytechnic University, Kowloon, Hong Kong\\
$^2$: Color, Imaging, and Metaverse Research Center, The Hong Kong Polytechnic University, Kowloon, Hong
Kong \\

\noindent \textit{\textbf{Team:}}\\
SCBC\\
\textit{\textbf{Members:}}\\
Lu Xu, Chao Zhang, Yasi Wang\\
\textit{\textbf{Affiliations:}}\\
SRC-Beijing\\

\noindent \textit{\textbf{Team:}}\\
OzU-VGL\\
\textit{\textbf{Members:}}\\
Furkan Kınlı$^{1,3,4}$, Doğa Yılmaz$^{1,3}$, Barış Özcan$^{2,3}$, Furkan Kıraç$^{1,3}$\\
\textit{\textbf{Affiliations:}}\\
$^1$: Vision and Graphics Laboratory, Özyeğin University, Türkiye\\
$^2$: Department of Computer Engineering, Bahçeşehir University, Türkiye\\
$^3$: Fishency Innovation, Norway \\
$^4$: T-Fashion Inc., Türkiye \& Canada \\

\noindent \textit{\textbf{Team:}}\\
MiAlgo\\
\textit{\textbf{Members:}}\\
Shuai Liu, Jingyuan Xiao, Chaoyu Feng, Hao Wang, Guangqi Shao, Yuqian Zhang, Yibin Huang, Wei Luo, Liming Wang, Xiaotao Wang, Lei Lei\\
\textit{\textbf{Affiliations:}}\\
Xiaomi Inc., China\\

\noindent \textit{\textbf{Team:}}\\
IVLTeam\\
\textit{\textbf{Members:}}\\
Simone Zini$^1$, Claudio Rota$^1$, Marco Buzzelli$^1$, Simone Bianco$^1$, Raimondo Schettini$^1$\\
\textit{\textbf{Affiliations:}}\\
$^1$: University of Milano - Bicocca, Italy\\

\noindent \textit{\textbf{Team:}}\\
DH-AISP\\
\textit{\textbf{Members:}}\\
Jin Guo, Tianli Liu, Mohao Wu, Ben Shao\\
\textit{\textbf{Affiliations:}}\\
Zhejiang Dahua Technology, China\\

\noindent \textit{\textbf{Team:}}\\
IIR-Lab\\
\textit{\textbf{Members:}}\\
Qirui Yang$^{1,2}$, Xianghui Li$^1$, Qihua Cheng$^{2}$, Fangpu Zhang$^1$, 
 Zhiqiang Xu$^2$, Jingyu Yang$^1$, Huanjing Yue$^1$\\
\textit{\textbf{Affiliations:}}\\
$^1$: School of Electrical and Information Engineering, Tianjin University, Tianjin 300072, China\\
$^2$: Shenzhen MicroBT Electronics Technology Co., Ltd, Shenzhen 518000, China \\

\section{Acknowledgements}

This work was supported by Artificial Intelligence Research Institute (AIRI). 
This work was partially supported by the Humboldt Foundation. We thank the NTIRE 2024 sponsors: Meta Reality Labs, OPPO, KuaiShou, Huawei and University of W\"urzburg (Computer Vision Lab).

{\small
\bibliographystyle{ieee_fullname}
\balance
\bibliography{egbib}

\begin{thebibliography}{10}\itemsep=-1pt

\bibitem{Pillow}
Pillow.
\newblock \url{GitHub - python-pillow/Pillow: Python Imaging Library (Fork)}.

\bibitem{abdelhamed2020ntire}
Abdelrahman Abdelhamed, Mahmoud Afifi, Radu Timofte, and Michael~S Brown.
\newblock {NTIRE} 2020 {C}hallenge on {R}eal {I}mage {D}enoising: {D}ataset,
  {M}ethods and {R}esults.
\newblock In {\em Proceedings of the IEEE/CVF Conference on Computer Vision and
  Pattern Recognition Workshops}, 2020.

\bibitem{ntire2024dehazing}
Cosmin Ancuti, Codruta~O Ancuti, Florin-Alexandru Vasluianu, Radu Timofte,
  et~al.
\newblock {NTIRE} 2024 dense and non-homogeneous dehazing challenge report.
\newblock In {\em Proceedings of the IEEE/CVF Conference on Computer Vision and
  Pattern Recognition (CVPR) Workshops}, 2024.

\bibitem{ashikhmin02}
Michael Ashikhmin.
\newblock {A} {T}one {M}apping {A}lgorithm for {H}igh {C}ontrast {I}mages.
\newblock pages 145--156, 01 2002.

\bibitem{banic2014improving}
Nikola Bani{\'c} and Sven Lon{\v{c}}ari{\'c}.
\newblock Improving the white patch method by subsampling.
\newblock In {\em 2014 IEEE International Conference on Image Processing
  (ICIP)}. IEEE, 2014.

\bibitem{banic2018flash}
Nikola Bani{\'{c}} and Sven Lon{\v{c}}ari{\'{c}}.
\newblock {F}lash and {S}torm: {F}ast and {H}ighly {P}ractical {T}one {M}apping
  based on {N}aka-{R}ushton {E}quation.
\newblock In {\em VISIGRAPP (4: VISAPP)}, pages 47--53, 2018.

\bibitem{Braun2006MemoryCE}
Karen~M Braun.
\newblock Memory color enhancement algorithm.
\newblock In {\em International Congress of Imaging Science}, 2006.

\bibitem{buades2005non}
Antoni Buades, Bartomeu Coll, and J-M Morel.
\newblock A non-local algorithm for image denoising.
\newblock In {\em 2005 IEEE Computer Society Conference on Computer Vision and
  Pattern Recognition (CVPR'05)}, volume~2. IEEE, 2005.

\bibitem{buchsbaum1980spatial}
Gershon Buchsbaum.
\newblock A spatial processor model for object colour perception.
\newblock {\em Journal of the Franklin Institute}, 310(1):1--26, 1980.

\bibitem{ntire2024QA_portrait}
Nicolas Chahine, Marcos~V. Conde, Sira Ferradans, Radu Timofte, et~al.
\newblock Deep portrait quality assessment. a {NTIRE} 2024 challenge survey.
\newblock In {\em Proceedings of the IEEE/CVF Conference on Computer Vision and
  Pattern Recognition (CVPR) Workshops}, 2024.

\bibitem{chen2018learning}
Chen Chen, Qifeng Chen, Jia Xu, and Vladlen Koltun.
\newblock {L}earning to {S}ee in the {D}ark.
\newblock In {\em CVPR}, 2018.

\bibitem{Chen_Chu_Zhang_Sun_2022}
Liangyu Chen, Xiaojie Chu, Xiangyu Zhang, and Jian Sun.
\newblock Simple baselines for image restoration.
\newblock Apr 2022.

\bibitem{ntire2024srx4}
Zheng Chen, Zongwei WU, Eduard~Sebastian Zamfir, Kai Zhang, Yulun Zhang, Radu
  Timofte, Xiaokang Yang, et~al.
\newblock {NTIRE} 2024 challenge on image super-resolution (×4): Methods and
  results.
\newblock In {\em Proceedings of the IEEE/CVF Conference on Computer Vision and
  Pattern Recognition (CVPR) Workshops}, 2024.

\bibitem{ntire2024rawsr}
Marcos~V. Conde, Florin-Alexandru Vasluianu, Radu Timofte, et~al.
\newblock Deep raw image super-resolution. a {NTIRE} 2024 challenge survey.
\newblock In {\em Proceedings of the IEEE/CVF Conference on Computer Vision and
  Pattern Recognition (CVPR) Workshops}, 2024.

\bibitem{mmseg2020}
MMSegmentation Contributors.
\newblock {MMSegmentation}: {O}pen{M}{M}{L}ab {S}emantic {S}egmentation
  {T}oolbox and {B}enchmark.
\newblock \url{https://github.com/open-mmlab/mmsegmentation}, 2020.

\bibitem{donoho1994ideal}
David~L Donoho and Iain~M Johnstone.
\newblock Ideal spatial adaptation by wavelet shrinkage.
\newblock {\em Biometrika}, 81(3):425--455, 1994.

\bibitem{nightimaging2022}
Egor Ershov et~al.
\newblock {NTIRE} 2022 {C}hallenge on {N}ight {P}hotography {R}endering.
\newblock In {\em Proceedings of the IEEE/CVF Conference on Computer Vision and
  Pattern Recognition (CVPR) Workshops}, 2022.

\bibitem{visapp24}
Egor Ershov, Artyom Panshin, Ivan Ermakov, Nikola Bani{\'{c}}, Alex Savchik,
  and Simone Bianco.
\newblock {R}eliability and {S}tability of {M}ean {O}pinion {S}core for {I}mage
  {A}esthetic {Q}uality {A}ssessment {O}btained {T}hrough {C}rowdsourcing.
\newblock In {\em Proceedings of the 19th International Joint Conference on
  Computer Vision, Imaging and Computer Graphics Theory and Applications -
  Volume 4: VISAPP}, pages 365--372. INSTICC, SciTePress, 2024.

\bibitem{getreuer2012rudin}
Pascal Getreuer.
\newblock {R}udin-{O}sher-{F}atemi {T}otal {V}ariation {D}enoising using
  {S}plit {B}regman.
\newblock {\em Image Processing On Line}, 2:74--95, 2012.

\bibitem{hu2017fc4}
Yuanming Hu, Baoyuan Wang, and Stephen Lin.
\newblock {FC4}: Fully convolutional color constancy with confidence-weighted
  pooling.
\newblock In {\em Proceedings of the IEEE Conference on Computer Vision and
  Pattern Recognition}, 2017.

\bibitem{ignatov2020aim}
Andrey Ignatov, Radu Timofte, Zhilu Zhang, Ming Liu, Haolin Wang, Wangmeng Zuo,
  Jiawei Zhang, Ruimao Zhang, Zhanglin Peng, Sijie Ren, et~al.
\newblock {AIM} 2020 challenge on learned image signal processing pipeline.
\newblock In {\em European Conference on Computer Vision}. Springer, 2020.

\bibitem{kang2010personalization}
Sing~Bing Kang, Ashish Kapoor, and Dani Lischinski.
\newblock Personalization of image enhancement.
\newblock In {\em 2010 IEEE Computer Society Conference on Computer Vision and
  Pattern Recognition}. IEEE, 2010.

\bibitem{kim2021large}
Dongyoung Kim, Jinwoo Kim, Seonghyeon Nam, Dongwoo Lee, Yeonkyung Lee, Nahyup
  Kang, Hyong-Euk Lee, ByungIn Yoo, Jae-Joon Han, and Seon~Joo Kim.
\newblock Large scale multi-illuminant (lsmi) dataset for developing white
  balance algorithm under mixed illumination.
\newblock In {\em Proceedings of the IEEE/CVF International Conference on
  Computer Vision}, pages 2410--2419, 2021.

\bibitem{ntire2024QA_UGC}
Xin Li, Kun Yuan, Yajing Pei, Yiting Lu, Ming Sun, Chao Zhou, Zhibo Chen, Radu
  Timofte, et~al.
\newblock {NTIRE} 2024 challenge on short-form {UGC} video quality assessment:
  Methods and results.
\newblock In {\em Proceedings of the IEEE/CVF Conference on Computer Vision and
  Pattern Recognition (CVPR) Workshops}, 2024.

\bibitem{ntire2024raim}
Jie Liang, Qiaosi Yi, Shuaizheng Liu, Lingchen Sun, Rongyuan Wu, Xindong Zhang,
  Hui Zeng, Radu Timofte, Lei Zhang, et~al.
\newblock {NTIRE} 2024 restore any image model ({RAIM}) in the wild challenge.
\newblock In {\em Proceedings of the IEEE/CVF Conference on Computer Vision and
  Pattern Recognition (CVPR) Workshops}, 2024.

\bibitem{liu2022deep}
Shuai Liu, Chaoyu Feng, Xiaotao Wang, Hao Wang, Ran Zhu, Yongqiang Li, and Lei
  Lei.
\newblock Deep-{F}lex{ISP}: {A} {T}hree-{S}tage {F}ramework for {N}ight
  {P}hotography {R}endering.
\newblock In {\em Proceedings of the IEEE/CVF Conference on Computer Vision and
  Pattern Recognition}, 2022.

\bibitem{ntire2024QA_AI}
Xiaohong Liu, Xiongkuo Min, Guangtao Zhai, Chunyi Li, Tengchuan Kou, Wei Sun,
  Haoning Wu, Yixuan Gao, Yuqin Cao, Zicheng Zhang, Xiele Wu, Radu Timofte,
  et~al.
\newblock {NTIRE} 2024 quality assessment of {AI}-generated content challenge.
\newblock In {\em Proceedings of the IEEE/CVF Conference on Computer Vision and
  Pattern Recognition (CVPR) Workshops}, 2024.

\bibitem{ntire2024lowlight}
Xiaoning Liu, Zongwei WU, Ao Li, Florin-Alexandru Vasluianu, Yulun Zhang,
  Shuhang Gu, Le Zhang, Ce Zhu, Radu Timofte, et~al.
\newblock {NTIRE} 2024 challenge on low light image enhancement: Methods and
  results.
\newblock In {\em Proceedings of the IEEE/CVF Conference on Computer Vision and
  Pattern Recognition (CVPR) Workshops}, 2024.

\bibitem{Menon2007DemosaicingWD}
Daniele Menon, Stefano Andriani, and Giancarlo Calvagno.
\newblock Demosaicing with directional filtering and a posteriori decision.
\newblock {\em IEEE Transactions on Image Processing}, 16:132--141, 2007.

\bibitem{moroney2000local}
Nathan Moroney.
\newblock Local color correction using non-linear masking.
\newblock In {\em Color and Imaging Conference}, volume 2000.1. Society for
  Imaging Science and Technology, 2000.

\bibitem{qian2019finding}
Yanlin Qian, Joni-Kristian Kamarainen, Jarno Nikkanen, and Jiri Matas.
\newblock On finding gray pixels.
\newblock In {\em Proceedings of the IEEE/CVF Conference on Computer Vision and
  Pattern Recognition}, 2019.

\bibitem{ntire2024efficientsr}
Bin Ren, Yawei Li, Nancy Mehta, Radu Timofte, et~al.
\newblock The ninth {NTIRE} 2024 efficient super-resolution challenge report.
\newblock In {\em Proceedings of the IEEE/CVF Conference on Computer Vision and
  Pattern Recognition (CVPR) Workshops}, 2024.

\bibitem{ronneberger2015u}
Olaf Ronneberger, Philipp Fischer, and Thomas Brox.
\newblock U-net: Convolutional networks for biomedical image segmentation.
\newblock In {\em International Conference on Medical Image Computing and
  Computer-assisted Intervention}. Springer, 2015.

\bibitem{schettini2010contrast}
Raimondo Schettini, Francesca Gasparini, Silvia Corchs, Fabrizio Marini,
  Alessandro Capra, and Alfio Castorina.
\newblock Contrast image correction method.
\newblock {\em Journal of Electronic Imaging}, 19(2):023005, 2010.

\bibitem{ntire2024shadow}
Florin-Alexandru Vasluianu, Tim Seizinger, Zhuyun Zhou, Zongwei WU, Cailian
  Chen, Radu Timofte, et~al.
\newblock {NTIRE} 2024 image shadow removal challenge report.
\newblock In {\em Proceedings of the IEEE/CVF Conference on Computer Vision and
  Pattern Recognition (CVPR) Workshops}, 2024.

\bibitem{wan2023seaformer}
Qiang Wan, Zilong Huang, Jiachen Lu, Gang Yu, and Li Zhang.
\newblock Seaformer: Squeeze-enhanced axial transformer for mobile semantic
  segmentation.
\newblock {\em arXiv preprint arXiv:2301.13156}, 2023.

\bibitem{ntire2024stereosr}
Longguang Wang, Yulan Guo, Juncheng Li, Hongda Liu, Yang Zhao, Yingqian Wang,
  Zhi Jin, Shuhang Gu, Radu Timofte, et~al.
\newblock {NTIRE} 2024 challenge on stereo image super-resolution: Methods and
  results.
\newblock In {\em Proceedings of the IEEE/CVF Conference on Computer Vision and
  Pattern Recognition (CVPR) Workshops}, 2024.

\bibitem{ntire2024lightfield}
Yingqian Wang, Zhengyu Liang, Qianyu Chen, Longguang Wang, Jungang Yang, Radu
  Timofte, Yulan Guo, et~al.
\newblock {NTIRE} 2024 challenge on light field image super-resolution: Methods
  and results.
\newblock In {\em Proceedings of the IEEE/CVF Conference on Computer Vision and
  Pattern Recognition (CVPR) Workshops}, 2024.

\bibitem{wang2004image}
Zhou Wang, Alan~C Bovik, Hamid~R Sheikh, and Eero~P Simoncelli.
\newblock Image quality assessment: from error visibility to structural
  similarity.
\newblock {\em IEEE Transactions on Image Processing}, 13(4):600--612, 2004.

\bibitem{ntire2024compressed}
Ren Yang, Radu Timofte, et~al.
\newblock {NTIRE} 2024 challenge on blind enhancement of compressed image:
  Methods and results.
\newblock In {\em Proceedings of the IEEE/CVF Conference on Computer Vision and
  Pattern Recognition (CVPR) Workshops}, 2024.

\bibitem{Yue:23}
Shuwei Yue and Minchen Wei.
\newblock Color constancy from a pure color view.
\newblock {\em J. Opt. Soc. Am. A}, 40(3):602--610, Mar 2023.

\bibitem{Yue:24}
Shuwei Yue and Minchen Wei.
\newblock Effective cross-sensor color constancy using a dual-mapping strategy.
\newblock {\em J. Opt. Soc. Am. A}, 41(2):329--337, Feb 2024.

\bibitem{ntire2024depth}
Pierluigi Zama~Ramirez, Fabio Tosi, Luigi Di~Stefano, Radu Timofte, Alex
  Costanzino, Matteo Poggi, et~al.
\newblock {NTIRE} 2024 challenge on {HR} depth from images of specular and
  transparent surfaces.
\newblock In {\em Proceedings of the IEEE/CVF Conference on Computer Vision and
  Pattern Recognition (CVPR) Workshops}, 2024.

\bibitem{zhang2018unreasonable}
Richard Zhang, Phillip Isola, Alexei~A Efros, Eli Shechtman, and Oliver Wang.
\newblock The unreasonable effectiveness of deep features as a perceptual
  metric.
\newblock In {\em Proceedings of the IEEE Conference on Computer Vision and
  Pattern Recognition}, 2018.

\bibitem{ntire2024bracketing}
Zhilu Zhang, Shuohao Zhang, Renlong Wu, Wangmeng Zuo, Radu Timofte, et~al.
\newblock {NTIRE} 2024 challenge on bracketing image restoration and
  enhancement: Datasets, methods and results.
\newblock In {\em Proceedings of the IEEE/CVF Conference on Computer Vision and
  Pattern Recognition (CVPR) Workshops}, 2024.

\bibitem{zhu2020metaiqa}
Hancheng Zhu, Leida Li, Jinjian Wu, Weisheng Dong, and Guangming Shi.
\newblock {M}eta{IQA}: {D}eep {M}eta-{L}earning for {N}o-{R}eference {I}mage
  {Q}uality {A}ssessment.
\newblock In {\em Proceedings of the IEEE/CVF Conference on Computer Vision and
  Pattern Recognition}, 2020.

\bibitem{zini2023back}
Simone Zini, Claudio Rota, Marco Buzzelli, Simone Bianco, and Raimondo
  Schettini.
\newblock Back to the future: a night photography rendering {ISP} without deep
  learning.
\newblock In {\em Proceedings of the IEEE/CVF Conference on Computer Vision and
  Pattern Recognition Workshops}, 2023.

\bibitem{zini2023shallow}
Simone Zini, Claudio Rota, Marco Buzzelli, Simone Bianco, and Raimondo
  Schettini.
\newblock {S}hallow {C}amera {P}ipeline for {N}ight {P}hotography
  {E}nhancement.
\newblock In {\em International Conference on Image Analysis and Processing},
  pages 51--61. Springer, 2023.

\end{thebibliography}
}

\end{document}